# A Bayesian Monte-Carlo Uncertainty Model for Assessment of Shear Stress Entropy

Amin Kazemian-Kale-Kale [a], Azadeh Gholami [a], Mohammad Rezaie-Balf [b], Amir Mosavi [c,d*],

Ahmed A Sattar [e], Bahram Gharabaghi [f], Hossein Bonakdari [g**]

[a] *Environmental Research Centre, Razi University, Kermanshah, Iran*

[b] *Department of Water Engineering, Graduate University of Advanced Technology, Kerman, Iran*

[c] *School of Built the Environment, Oxford Brookes University, Oxford OX30BP, UK;*

*a.mosavi@brookes.ac.uk*

[d] *Institute of Structural Mechanics, Bauhaus University Weimar, D-99423 Weimar, Germany;*

*amir.mosavi@uni-weimar.de*

[e] *Department of Irrigation & Hydraulics, Faculty of Engineering, Cairo University, Giza, German University in Cairo, Egypt*

[f] *School of Engineering, University of Guelph, Guelph, Ontario, NIG 2W1, Canada*

[g] *Department of Soils and Agri-Food Engineering, Laval University, Québec, Canada, G1V0A6;*

*hossein.bonakdari@fsaa.ulaval.ca*

**Abstract:** The entropy models have been recently adopted in many studies to evaluate the distribution of the shear stress in circular channels. However, the uncertainty in their predictions and their reliability remains an open question. We present a novel method to evaluate the uncertainty of four popular entropy models, including Shannon, Shannon-Power Low (PL), Tsallis, and Renyi, in shear stress estimation in circular channels. The Bayesian Monte-Carlo (BMC) uncertainty method is simplified considering a 95% Confidence Bound (CB). We developed a new statistic index called as $FREE_{opt}$-based OCB ($FOCB$) using the statistical indices Forecasting Range of Error Estimation (FREE) and the percentage of observed data in the CB ($N_{in}$), which integrates their combined effect. The Shannon and Shannon PL entropies had close values of the $FOCB$ equal to 8.781 and 9.808, respectively, had the highest certainty in the calculation of shear stress values in circular channels followed by traditional uniform flow shear stress and Tsallis models with close values of 14.491 and 14.895, respectively. However, Renyi entropy with much higher values of $FOCB$ equal to 57.726 has less certainty in the estimation of shear stress than other models. Using the presented results in this study, the amount of confidence in entropy methods in the calculation of shear stress to design and implement different types of open channels and their stability is determined.

# 1. Introduction

When studying sediment transport of sewer pipes and scour of the deposited materials or the channel walls, accurate calculation of boundary shear stress distribution is very important. At many places within urban sanitary or storm sewer networks, the flow tends to slow down, allowing carried sediment to be deposited on channel bed. With time and continuous deposition of sediment, the circular bed shape tends to build up in the form of irregular or flatbeds. This leads to the change in the channel shape, which alters the hydraulic characteristics of flow and shear distribution in the channel. This process is often encountered in practice, and many circular channels exhibit flatbeds. In circular open channels, the shear stress distribution has been experimentally analyzed by many researchers [1-3]. Others have developed models to predict the distribution of shear stress along a channel using numerical and analytical methods [4-6]. Numerical models depending on soft computing (SC) techniques have been recently presented for the prediction of shear stress distribution [7,8].

The entropy concept has been used by researchers to predict different flow variables such as velocity distribution [9-11], transverse slope of stable channels banks [12-14], and shear stress distribution [15]. Chiu [16], for the first time, introduced models for calcualting the shear stress in open channels using the Shannon entropy concept. Sterling and Knight [17] developed equations based on Shannon entropy for estimating boundary shear stress in open channels. Although, their results showed this approach could estimate the distribution of the shear stress reasonably well, however, they indicated some practical limitations in application of the Shannon entropy for practical cases. Sheikh and Bonakdari [18] employed the entropy concept and proposed Power Law (PL) techniques to develop new equations for predicting shear stress distribution in open channels. Their results compared with measured data and showed that their proposed Shannon PL model had good potential for practical applications besides Shannon entropy.

Other entropies such as Tsallis and Renyi concepts can be applied to measure the probability of variables [14,19]. Tsallis [20] proposed Tsallis entropy as a generalization of the Shannon entropy comprising a supplementary parameter. Renyi and Tsallis entropies with non-additive parameters can be utilized to perform it less susceptible to the form of the probability distribution [14, 21-23]. Bonakdari et al. [24] showed that the Tsallis method could calculate the shear stress distribution along the wetted perimeter with reasonable accuracy. Khozani and Bonakdari [25] employed the Renyi entropy and presented a relationship for estimation of the shear stress distribution in circular channels. They concluded that their presented model showed the ability to calculate the shear stress distribution and compared well to measured results. Although many studies carried out on entropy models for prediction of shear stress distribution, wide implementation of entropy models has not taken place due to the absence of enough confidence in these models compared to previous conventional models in estimation of flow variables. Thus, it is beneficial to analyze the uncertainty of entropy models and compare the performance of them with previous conventional models.

The uncertainty analysis of many hydraulic and hydrological models has been tackled by many researchers using various uncertainty methods [26-40]. Lamb et al. [41] used measured water depth in a Bayesian procedure as Generalized Likelihood Uncertainty Estimation (GLUE) to measure uncertainties for a simple model of the rainfall-runoff distribution. Misirli et al. [36] used the Bayesian Recursive Estimation (BaRE) to evaluate the watershed model's uncertainty. They introduced the *FREE* statistic index for evaluating models' uncertainty. *FREE* is equal to the absolute sum of the measured data within the confidence bound ($F_P$) and the absolute sum of the measured data outside the confidence bound ($F_N$). Thus, the *FREE* index represents the width of the CB. Corato et al. [42], using the uncertainty method introduced by Misirli et al. [36], analyzed the uncertainty of the Shannon entropy model in estimating the velocity distribution proposed by Moramarco et al. [43]. The proposed model for estimating velocity in their study was satisfactory for calculating the average velocity in multiple segments with 95% CB, but they did not present any results for the model's ability to calculate the velocity distribution at the channel cross-section. Rincon [44] provided a comprehensive overview of uncertainty methods that are widely used in hydrodynamics. Rincon [44] concluded that BMC is a powerful method for evaluating sources of uncertainty in hydrodynamic modeling. He stated that despite the BMC calculations required, this method could be easily applied in most cases.

In the fields of measuring the uncertainty of entropy models in the estimation of shear stress distributions in open channels, based on the knowledge of authors, no study is seen except the recent studies related to authors. Kazemian-Kale-Kale et al. [45], analyzed the Tsallis entropy uncertainty in the prediction of shear stress distribution in circular channels for the first time. Their proposed uncertainty method was based on models with Gaussian (normal) error distributions. Therefore, firstly, they studied the error distribution of shear stress predicted by Tsallis entropy and they emphasized on the necessity to normalize the shear stress data for the uncertainty analyses. Then, they calibrated the model to select the best sample size (shear stress data considered under different hydraulic conditions) and finally analyzed the uncertainty of the Tsallis entropy model using this Sample Size (SS). Although their results were well capable of evaluating the uncertainty of the Tsallis model, the calibration method was difficult, and in addition, they did not discuss the performance of the different transfer functions to normalization in the uncertainty results. Kazemian-Kale-Kale et al. [46] also analyzed the uncertainty of

Shannon entropy in calculating shear stress in a circular channel. They simplified the calibration method of Kazemian-Kale-Kale et al. [45], and Johnson's function was employed to analyze the Shannon entropy uncertainty. Kazemian-Kale-Kale et al. [45, 46] studied the uncertainty of the Tsallis and Shannon entropy models in calculating of shear stress distribution. However, because these statistics need to be examined simultaneously to determine each model's uncertainty, they were not able to compare the uncertainty of different entropy models. Therefore, they did not discuss the reliability of the Shannon and Tsallis entropy models in calculating shear stress compared to other models.

Based on these gaps, this paper aims at providing a novel uncertainty model for quantifying and evaluating the uncertainty of four different entropy models; Shannon, Shannon PL, Tsallis, and Renyi, in calculating shear stress in the circular channels. The obtained results are compared to those calculated by the established analytical equation for shear stress in uniform flow. The uncertainty method introduced by Kazemian-Kale-Kale et al. [46] based on the BMC method is briefly presented as the Hybrid Bayesian and Monte-Carlo Estimation System (HBMES-1) then further improved, and the uncertainty method is presented in this study as HBMES-2. Using the improved model, the uncertainty of several different entropy models and the optimized/minimum CB that covers all measured data (*OCB*) are defined. The improved model is based on the optimized OCB (width of the confidence bound) and the absolute sum of measured data $FREE_{opt}$. Given the value of *OCB* and $FREE_{opt}$, the *FOCB* statistic is introduced that can show the effect of all uncertainty statistics. *FOCB* shows the degree of qualitative and quantitative uncertainty of shear stress models. Since the main purpose of current study is to assess and compare the uncertainty of four models in prediction of shear stress, it is easy to make a comparison using

HBMES-2 results, due to the using only one statistic in HBMES-2 method, the accuracy of the models could be determined.

## 2. Material and methods

### 2.1. Entropy models

#### 2.1.1. Shannon model

Sterling and Knight [17] employed Shannon entropy to present a model to predict shear stress distribution in circular channels as follow:

$$\tau = \frac{1}{\lambda_0} \ln\left[1 + (e^{\lambda_0 \tau_{max}} - 1)\frac{y}{P/2}\right] \tag{1}$$

where $\tau$ is the local shear stress distribution, $\tau_{max}$ is the maximum shear stress, $P$ is the wetted perimeter of the circular channel, and $y$ is the specific point in wetted perimeter, which we want to obtain the shear stress on it; value changes between 0 to $P/2$. Figure 1 shows the circular cross-section with related notations used in models. The shear stress values are only computed for half the wetted perimeter due to section symmetry and, consequently, hydraulic characteristics. In Equation (1) $\lambda_0$ is the Lagrange multiplier that can be determined as:

$$\frac{1}{\lambda_0} = \left[\frac{\tau_{max} e^{\lambda_0 \tau_{max}}}{e^{\lambda_0 \tau_{max}} - 1} - \rho g R s\right]^{-1} \tag{2}$$

where $\rho$ is the fluid density, $g$ is the gravity coefficient, $R$ is the hydraulic radius, and $s$ is the channel slope. In a circular channel with a flatbed, Equation (1) can be written as follows:

$$\tau_w = \frac{1}{\lambda_{0w}}\left[1 + \left(e^{\lambda_{0w}\tau_{max(w)}} - 1\right)\frac{2(y - y_w)}{P_w}\right] \qquad y_w \leq y \leq \frac{P_w}{2} \tag{3}$$

$$\tau_b = \frac{1}{\lambda_{0b}}\left[1+\left(e^{\lambda_{0b}\tau_{\max(b)}}-1\right)\frac{2(y-y_w)}{P_b}\right] \qquad \frac{P_w}{2}\le y \le \frac{P_b}{2}+y_w \tag{4}$$

where $\tau_{\max(w)}$ and $\tau_{\max(b)}$, are the maximum shear stress at the wall and bed of channel respectively, $P_w$ and $P_b$ are the wetted perimeter corresponding to the wall and bed of the channel, and $y_w$ is an offset and taken as 5 mm in their analysis.

The equations presented by Knight et al. [47] were used to predict the mean and maximum shear stress for wall and bed of circular with the flatbed channel as:

$$\frac{\bar{\tau}_w}{\rho g R S} = 0.01\% SF_w\left(1+P_b/P_w\right) \tag{5}$$

$$\frac{\bar{\tau}_b}{\rho g R S} = \left(1-0.01\% SF_w\right)\left(1+P_b/P_w\right) \tag{6}$$

$$\frac{\tau_{\max(w)}}{\rho g R S} = 0.01\% SF_w\left[2.0372(P_b/P_w)^{0.7108}\right] \tag{7}$$

$$\frac{\tau_{\max(b)}}{\rho g R S} = \left(1-0.01\% SF_w\right)\left[2.1697(P_b/P_w)^{-0.3287}\right] \tag{8}$$

where $\bar{\tau}_w$ and $\bar{\tau}_b$ are the mean wall, and bed shear stress, respectively, and $\%SF$ is the wall shear force percentage determined by the following equation:

$$\% SF_w = C_{sf}\exp\left(-3.23\log(P_b/C_2 P_w+1)+4.6052\right) \tag{9}$$

where $C_2 = 1.38$ and $C_{sf} = 1.0$ for $\frac{P_b}{P_w} < 4.374$, otherwise $C_{sf} = 0.6603(P_b/P_w)^{0.28125}$.

**Figure 1.** Cross-section of the circular with the flatbed channel and its notation.

### 2.1.2. Shannon PL model

Sheikh and Bonakdari [18] proposed the following equation to predict shear stress in circular channels.

$$\tau = \tau_{max}(\frac{y}{P/2})^{1/n} \tag{10}$$

where $n$ is a non-dimensional parameter computed as:

$$n = \frac{\bar{\tau}}{\tau_{max} - \bar{\tau}} \tag{11}$$

where $\bar{\tau}$ is the mean shear stress value. It should be noted that these equations are used to predict the wall and bed sear stress of circular with flatbeds channel separately. The mean and maximum shear stress values are given in Equation (11) are obtained from Equations (5)-(8). In the entropy models presented below, the Equations (5)-(8) are also used.

### 2.1.3. Tsallis entropy model

Bonakdari et al. [24] employed the concept of Tsallis entropy to present the following relationship for estimating shear stress in a circular channel:

$$\tau = \frac{k}{\lambda_1}\left[(\frac{\lambda_2}{k})^k + \frac{\lambda_1 y}{P}\right]^{1/k} - \frac{\lambda_2}{\lambda_1} \tag{12}$$

where $k = \frac{q-1}{q}$, $q$ is a real value, $\lambda_1$ and $\lambda_2$ are Lagrange multipliers that can be determined from the two following equations:

$$[\lambda_2 + \lambda_1 \tau_{max}]^k - [\lambda_2]^k = \lambda_1 k^k \tag{13}$$

$$\tau_{max}(k+1)\lambda_1[\lambda_2 + \lambda_1 \tau_{max}]^k - [\lambda_2 + \lambda_1 \tau_{max}] = (k+1)\lambda_1^2 k^k \bar{\tau} \tag{14}$$

### 2.1.4. Renyi entropy model

Khozani and Bonakdari [25] employed Renyi entropy to estimate the distribution of the shear stress distribution and introduced the following equation:

$$\tau = \tau_{max}\left(\frac{1}{\lambda'}\left[-\lambda'' - \left((-\lambda'')^{k'} - \frac{\lambda'\alpha'^{k'}}{(\alpha'-1)}\frac{y}{P/2}\right)^{1/k'}\right]\right) \quad (15)$$

where $k' = \frac{\alpha'}{\alpha'-1}$ and $\alpha'$ is a real number between zero and one. $\lambda'$ and $\lambda''$ are Lagrange multipliers that can be calculated with two following equations [25]:

$$\frac{(-\lambda''-\lambda')^{k'} - (-\lambda'')^{k'}}{\lambda'} = \frac{\alpha'^{k'}}{1-\alpha'} \quad (16)$$

$$\frac{-1}{\lambda'}(-\lambda''-\lambda')^{k'} - \frac{1}{\lambda'^2(k'+1)}\left[(-\lambda''-\lambda')^{k'+1} - (-\lambda'')^{k'+1}\right] = \frac{\alpha'^{k'}}{(\alpha'-1)}\hat{\bar{\tau}} \quad (17)$$

where $\hat{\bar{\tau}} = \bar{\tau}/\tau_{max}$ is the dimensionless mean shear stress.

## 2.2. Global shear stress (ρgRs)

The shear stress in open channels in case of uniform flow is considered as a basic model for comparison with entropy models as follows:

$$\tau = \rho g R s \quad (18)$$

## 2.3. Data Collection

The measured shear stress data were collected from experimental results presented by Sterling [48]. The author measured the values of shear stress along the wetted perimeter of a circular channel with a diameter of 244mm in different flow conditions, as shown in Table 1. For simulating a circular with a flatbed channel, Sterling [48] laid a thickness of sediment in a circular channel. As seen in Table 1, the values of the shear stress are measured in four flow depths in the circular channel ($t/D = 0$) and the rest of values related to circular with a flatbed channel ($t/D \neq 0$) with different bed-sediment thickness. These measured values shall be considered in the uncertainty analyses of the four entropy models.

**Table 1**. Summary of the hydraulic parameters in the circular channel with and without sediment.

## 2.4. Uncertainty analysis

The uncertainty method presented in Kazemian-Kale-Kale et al. [46] is develpoed to analyze the uncertainty of four entropy models of shear stress predictor. This method is named HBMES-1 (Hybrid Bayesian and Monte-Carlo Estimation System in the first stage). The HBMES-1 uncertainty is improved and will be used to analyze the four entropy models in shear stress prediction. This method of uncertainty will be described further in this study as HBMES-2.

## 2.4.1. HBMES-1 uncertainty method

The basis of the uncertainty method presented by Kazemian-Kale-Kale et al. [45,46] was that the error distributions of the understudy models were assumed to follow a Gaussian (normal) distribution. Due to the importance of normalizing the data for the uncertainty analysis in the study of Kazemian-Kale-Kale et al. [46], two common transfer functions of Box-Cox and Johnson were compared to evaluate the shear stress estimations models' uncertainty. Kazemian-Kale-Kale et al. [45] considered the 95% CB and performed 15 tests for uncertainty calibration based on the same CB, considering shear stress data under different hydraulic conditions.

These tests were performed to select the best sample size (shear stress data considered under different hydraulic conditions). In order to choose the best SS (sample size), they examined the variation of the $N_{in}$ mean and the Box-Cox function transfer factor. Due to the combination of these two statistics, the best SS based on the method of Kazemian-Kale-Kale et al. [45] will be difficult, especially when the uncertainty of several models is considered.

In this study, the best SS is selected based on the mean value of $N_{in}$. According to Corato et al. [42], the closer $N_{in}$ is to 95% CB, the assumption of the Gaussian error distribution is more satisfying. Considering $N_{in}$'s changes, satisfy the initial condition of the Gaussian error distribution. After selecting the best SS, the best transfer factor value is used for the final evaluation of the uncertainty. The results of this evaluation are in multiple statistics of $N_{in}$, $F_P$, $F_N$, and *FREE*. Using these statistics, it is determined whether each of the entropy models is sufficiently certainty to predict shear stress.

### 2.4.2. HBMES-2 uncertainty method

As the HBMES-1 method requires multiple statistics to be evaluated concurrently, comparing the certainty of several models using this method is very difficult and in some cases are impossible. In this study, the HBMES-1 uncertainty method has been improved, and the result of uncertainty for each model is presented as a single statistic.

The improved method in this paper called HBMES (in second stage) or HBMES-2. The HBMES-2 uncertainty process is in two parts: (1) The calibration section, and (2) The final analysis section and the introduction of new statistics. The calibration part of the HBMES-2 method is similar to the HBMES-1 method and is performed with the 95% CB to select the best SS [46]. For the final analysis, a minimum CB that covers all shear stress measured data is found. This minimum CB is called *OCB* (Optimized Confidence Bound). The HBMES-2 uncertainty analysis process is described using the following three steps.

*Step 1. Determining initial borders of the OCB ($OCB_i$)*

First, the shear stress data are normalized using the Box-Cox transfer function and transfer factor obtained in the best SS, and then the Gaussian error distribution is calculated as follows:

$$\varepsilon(y/P) = Z_m(y/P) - Z_p(y/P) \tag{19}$$

where $\varepsilon(y/P)$ is the error of data normalization, $Z_e$, and $Z_p$ are the normalized values of measured and predicted shear stress ($\tau_m$ and $\tau_p$), respectively.

Then with considering a given value for $OCB_i$, the following relation is applied:

$$(Z(y/P)|^{\pm})_i = Z_p(y/P) + \mu_\varepsilon \pm (\alpha/2)_i \delta_\varepsilon \tag{20}$$

where I = 0, 1, 2, ..., n. This value, $(Z|^{\pm})_i$, indicates the effects of the considered $OCB_i$ and the Gaussian error distribution (Equation 19) on the shear stress data predicted by the model. $(\alpha/2)_i$ is the standard normal curve variable, which is related to the percentage of considered $OCB_i$.

By performing this dividing operation;

$$\frac{(OBC)_i/100}{2} = \psi_i \tag{21}$$

where $\psi_i$ is the area below one side of the probability distribution diagram. Using $\psi_i$ from the standard normal curve table, the value of the corresponding standard coefficient $(\alpha/2)_i$ is obtained. The first assumption for $OCB_i$ $(OCB)_0 = 95\%$, which will be $\psi_0 = 0.4750$ by the above standard deviation from the table of the corresponding standard coefficient $(\alpha/2)_0 = 1.96$. In the Equation (20), $\mu_\varepsilon$ and $\delta_\varepsilon$ are the mean and standard deviation of the Gaussian error distribution of the normalized shear stress data, respectively. Then the $OCB_i$ borders are obtained by the Box-Cox transfer function through the following relation:

$$(\tau(y/P)|^{\pm})_i = \begin{cases} [(Z(y/P)|^{\pm})_i \lambda + 1]^{1/\lambda} & if \quad \lambda \neq 0 \\ \exp[(Z(y/P)|^{\pm})_i] & if \quad \lambda = 0 \end{cases} \tag{22}$$

where $(\tau|^{\pm})_i$ are the borders of $OCB_i$ and $\lambda$ is the Box-Cox transfer factor.

After determining $OCB_i$ borders, the distance of the measured data from the borders ($dist_x$) and the sum of the distances (*FREE*) using the equations of Kazemian-Kale-Kale et al. [46] can be calculated. If the $dist_x$ value is positive, then the data is inside $OCB_i$ and if the $dist_x$ is negative, then the data is outside $OCB_i$. As a result, the value of $N_{in}$ (the percentage of measured data within the $OCB_i$) using $dist_x$ values is obtained.

*Step 2. Assessment of the final OCB ($OCB_n$)*

Given that in the $OCB_n$, all points are within the CB, $N_{in}$ should always be equal to 100%. If $N_{in}$ is equal to 100, the $OCB_i$ value is considered lower, and the three steps a-c are done. However, if $N_{in}$ is less than 100, the $OCB_i$ values considered higher, and the three steps a-c is performed. This should continue until $N_{in}$ is equal to 100% and the *FREE* values are at their lowest value. This *FREE* value is called $FREE_{opt}$. How to determine $OCB_n$ is shown in Figure 2. In this figure, $\xi$ is the accuracy of the standard normal table and equal to $10^{-2}$.

**Figure 2.** Process of assessment of the $OCB_n$ (minimum confidence bound that covers all measured data) in HBMES -2 uncertainty method.

*Step 3. Introducing uncertainty index of FOCB*

After determining $OCB_n$, the final borders of $(\tau_n/^{\pm})_n$ are obtained from Equation (22). In $OCB_n$ the $dist_x$ value presented in the study by Kazemian-Kale-Kale et al. [46] is optimized and derived from the following equation:

$$dist_{x(opt)} = \begin{cases} (\tau|^+)_n - \tau|_m & if \quad \tau|_m - \tau|_p \geq 0 \\ \tau|_m - (\tau|^-)_n & if \quad \tau|_m - \tau|_p < 0 \end{cases} \quad (23)$$

where $\tau_m$ and $\tau_p$ are the values of measured and predicted shear stress, respectively. $(\tau|^{\pm})_n$ assesses the borders of $OCB_n$, which is obtained by inserting the variable of standard normal curve that relates to $OCB_n$ as $(\alpha/2)_n$ (Equation 22). With considering $OCB_n$, the values of $FREE$ [46] are always positive, and in this paper is called $FREE_{opt}$ that written as follows:

$$FREE_{opt} = F_P = \sum_{dist_x > 0} dist_x \quad (24)$$

$FREE_{opt}$ index equals to the sum of the intervals of measured values inside the $OCB_n$. The $OCB_n$ and $FREE_{opt}$ statistics are quantitative and qualitative criterion, respectively that shall be used to examine the uncertainty of the four shear stress predictor models. To consider the combined effect of these two statistics, the $FOCB$ statistic is introduced such as that:

$$FOCB = \frac{OCB_n * FREE_{opt}}{100} \quad (25)$$

Given that $OCB_n$ is used as a result, the present research uses the term $OCB$ for convenience.

## 3. Result and discussion

Since the calibration of both the uncertainty analysis methods presented in Sections 2.4.1 and 2.4.2 are performed the same, the results of the uncertainty calibration of the four entropy models are shown first. Then, the results of the HBMES-1 uncertainty performed at the $CB = 95\%$ and the HBMES-2 results obtained with $OCB$ are presented.

### 3.1. Calibration

As mentioned, the calibration was done based on $N_{in}$ changes to select the best SS. 15 tests were performed for calibration, and the difference in each test varied in the number of data considered under hydraulic conditions. To select the best SS in each model, $N_{in}$ values of each test (SS = 9 to SS = 23) were obtained at the calibration step with a default CB of 95%. Figure 3 illustrates the changes in $N_{in}$ values as a box plot for all four entropy models. As stated before, each SS that has an average $N_{in}$ value is closer to 95%, is selected as the best SS. As seen in the figure, for the Shannon, Shannon PL, Tsallis, and Renyi entropies, the closest $N_{in}$ values to the dotted line (95%) are 94.85%, 95%, 94.79%, and 94.23%, respectively. These values occur in SS amounts of 17, 15, 18, and 13, respectively. Therefore, the obtained SS values for each entropy model based on $N_{in}$ values can be introduced as the best SS values in the calibration phase. After selecting the best SS, the value of Box-Cox transfer factor ($\lambda$) in this SS is considered for the final evaluation of the uncertainty of each model. So, the values of transfer factor ($\lambda$) are calculated for all models in different values of SS. These values of the $\lambda$ include the average, upper, and lower limits for each SS (according to Figure 4). As seen in Figure 4, CB changes in all models are not more significant than the mean values of the $\lambda$. Thus, the mean amount of the $\lambda$ in the best SS (which is chosen based on $N_{in}$ values) is selected as the best $\lambda$. According to this figure, all model trends are similar and ascending. Changes of CB and also $\lambda$ values in the initial values of SS are high, and with increasing SS value, the $\lambda$ values decrease. The value of the $\lambda$ in the optimized SS in the previous section (Figure 3) is determined as the best transfer factor ($\lambda$).

**Figure 3.** The changes in the percentage of measured values within the confidence bound ($N_{in}$) in different SS values for all entropy models.

According to Figure 4, for all entropy models, after reaching the best SS value, the variation in the $\lambda$ is very low and tends to become constant. The constant changes of the factor $\lambda$ (zero change) after the best SS indicates that in order to achieve a Gaussian error distribution in the shear stress data series, the minimum test for each model is in the best SS value. Moreover, the notable point is the similar trend in $\lambda$ graphs and its change in three entropies of Tsallis, Shannon, and Shannon PL model, unlike the Renyi entropy model. Unlike others, in Renyi entropy, the CB variations are so significant in all SS values. Therefore, the less reliability of Renyi entropy is evidence and clear in here.

**Figure 4.** Changes of transfer factor ($\lambda$) values in different sample sizes for all entropy models.

The validity of the uncertainty analysis results depends on the normality of the error distribution of each entropy model. The mean and standard deviation of the Gaussian error distribution ($\delta_\varepsilon$ and $\mu_\varepsilon$) are also examined in the calibration stage along with the selection of the best SS. Hence, the changes of $\lambda$ were obtained for various SS values $\delta_\varepsilon$ and $\mu_\varepsilon$ were calculated for different SS (sample size) values. Using the $\delta_\varepsilon$ statistic, the degree of compliance of the shear stress error distribution obtained from each entropy is determined from the Gaussian distribution. The closer the error distribution is to the Gaussian distribution, leads to the greater the validity of the uncertainty results.

In Figure 5, the variations of $\delta_\varepsilon$ for different SS values are illustrated for the four entropy models. As observed in these Figures, for all models, the mean value of $\delta_\varepsilon$ shows the constant trend in different SS values, and these values are 0.09, 0.06, 0.08, and 0.21 for the Shannon, Shannon PL, Tsallis and Renyi entropies, respectively. The lower the value of $\delta_\varepsilon$, the more the error distribution of the transferred data will follow Gaussian distribution. Therefore, due to the small values of $\delta_\varepsilon$ for the error distribution of the shear stress obtained from the Shannon, Shannon PL, and Tsallis, compared to the Renyi entropy, it can be said that the error distribution of these three models compared to the Renyi model are closer to the Gaussian distribution. Also, the Renyi entropy with the $\delta_\varepsilon$ value of three times larger than those of others did not have a favorable outcome.

The variation of the $\mu_\varepsilon$ value is derived from all four entropy models, as illustrated in Figure 6. As shown in this figure, the changes of $\mu_\varepsilon$ for the Renyi entropy model is higher than those of the Shannon, Shannon PL, and Tsallis models. The $\mu_\varepsilon$ values for all four models in the different SS have an almost constant value, but because the range of $\mu_\varepsilon$ variations for Renyi entropy is much higher than other models so that this value of $\mu_\varepsilon$ is less reliable in this model. The absolute values of average $\mu_\varepsilon$ for Shannon PL, Tsallis, Shannon, and Renyi models are 0.001, 0.003, 0.055, and 0.114, respectively. Therefore, the two entropies of Shannon PL and Tsallis contain less error than the Shannon entropy, and these three entropies perform better than Renyi entropy.

**Figure 5.** The mean and standard deviation of the Gaussian error distribution; $\delta_\varepsilon$ and $\mu_\varepsilon$, respectively; versus sample size in all entropy models.

*3.2. Assessment of uncertainty of four entropy models using the HBMES-1 method*

By plotting the 95% CB using the HBMES-1 method introduced in Section 2.4.1, the uncertainty statistics for the 4 entropy models and the global shear stress model introduced in Section 2.2 were obtained and presented in Table 2. According to the researchers' studies, the predictor model is highly reliable if 80-100% of the values are in the desired CB and the model was not able to predict if less than 50% of the values are within the CB [30,32,49,50].

Because of these studies, given the percentages of measured shear stress data within the CB equal to 95% ($N_{in}$), it can generally be said whether the entropy models are sufficiently accurate in estimating shear stress or not?. The values of $N_{in}$ in column 2 of Table 2 represent the required certainty for all models in estimating shear stress since the $N_{in}$ value for all models is greater than 85%. $N_{in}$ values in all entropy models are very close together and higher than $N_{in}$ values for the conventional $\rho gRs$ model.

Although $N_{in}$ values can be used for certainty, to compare each model with another, $F_P$, $F_N$, and *FREE* values should be considered in addition to $N_{in}$ values. The $F_P$ values given in the third column of Table 2 represent the absolute sum of the internal data from the borders of CB. The $F_N$ values given in the fourth column of Table 2 represent the absolute sum of the outer data from the borders of CB. The *FREE* values, which are equal to the sum of $F_P$ and $F_N$, represent the Width of Confidence Bound (WCB) in the last column of Table 2.

The lower these *FREE*, $F_P$, and $F_N$ values, means the higher model's certainty. These three values are close for the two Shannon and Shannon PL entropies as well as for the two Tsallis and $\rho gRs$ models, but these values are much higher for the Renyi entropy than the other four models. As can be seen, the $N_{in}$ values with the values of the three $F_P$, $F_N$, and *FREE* statistics give different results in providing the uncertainty, and it is difficult to give a precise view to compare the accuracy of the models. For example, although the $N_{in}$ value for the Renyi entropy is more than $ρgRs$, the $F_P$, $F_N$ and $FREE$ values are much better (lower) for the $ρgRs$ model.

Therefore, the general conclusion from the values in this table is that given the high values of $N_{in}$, all models are capable of predicting shear stress with high precision, but an accurate comparison of this uncertainty with respect to other statistics to be considered concurrently is complicated. In this study, HBMES-2 method is presented to solve this problem. In the next section, the results are provided and discussed in details.

**Table 2.** Statistical indexes based on HBMES-1 uncertainty method in shear stress prediction by different entropy models and conventional $ρgRs$ model.

To clarify the results presented in Table 2, the shear stress distribution and 95% CB using the HBMES-1 method for a height ratio of the circular channel (t/D=0, h+t/D=0.333) and the height ratio of the circular channel with flatbed (t/D=0.25, h+t/D=0.333) is shown in Figure 6. It is seen in the figure that the trend of shear stress distribution in all four entropy models is in line with the trend of measured values for the bed area and the channel walls, whereas the conventional $ρgRs$ model has both a constant amount in the bed and the walls of the channel.

In the circular channel (Figure 6a), the entropy models of Shannon, Shannon PL, and Tsallis have predicted the shear stress distribution quite in accordance with the measured data. The Renyi entropy is also in good agreement with measured data except for channel sides (0<y/P<0.1 and 0.9<y/P<1). However, the performance of the Renyi entropy is much better than the $ρgRs$. The $ρgRs$ model at the walls of the circular channel predicts the shear stress value much lower than the measured values, and at the bed area (0.1<y/P<0.9), it predicts the shear stress value slightly below the corresponding measured values.

Consequently, with designing, the channel based on the $\rho gRs$ model, both the resistance of the walls is considered unnecessarily high, and the scouring of the bed also occurs. In the circular with flatbed channel (Figure 6b), the results of all models are similar to the circular channel. However, the Shannon PL entropy, which is in accordance with the measured trend, predicts the shear stress values more than the two Shannon and Tsallis entropies.

To check the uncertainty of the models, the WCB and the percentage of measured data within the CB ($N_{in}$) in Figures 6 should be considered. The $N_{in}$ value and the WCB at these height ratios were calculated according to the overall ability of the entropy models and considering all the 23 different hydraulic conditions in Table 1. Given that the $N_{in}$ in both height ratios for all models is more than 93%, all models are capable of estimating shear stress with high certainty. Figure 6a shows that the CB for the Shannon and Shannon PL models with all data covered is very small and almost uniform.

In the Tsallis and Renyi models, two shear stress data in free surface (y/P=0) fall outside the CB (hollow points in Figure 6), indicating that the certainty of these models in predicting shear stress at the free surface of the circular channel is less than the other wet areas. This issue is more seen by Renyi entropy model. On the other hand, the WCB of the Tsallis model is much less than the WCB of the Renyi model, which demonstrates the more reliability in addition to the accuracy of Tsallis model especially in areas near the water surface.

The WCB of the $\rho gRs$, Shannon, and Shannon PL, covered all data, but the WCB in the $\rho gRs$ is larger than the other two models. Furthermore, the $\rho gRs$ model is not capable of estimating shear stress values; so that there is no consistency with observed values. When comparing the certainty of the $\rho gRs$ and Tsallis model, it should be noted that although the Tsallis entropy has two data at the free surface outside the CB, the WCB is much lower in the Tsallis model than the $\rho gRs$ model.

Therefore, the certainty of Tsallis model is much more than $\rho gRs$ model in addition to more accuracy of the Tsallis model. The Renyi model also has the lowest certainty in predicting shear stress at this height ratio with the highest WCB and the least amount of $N_{in}$. It is also observed in Figure 6b that the WCB for the three Shannon, Shannon PL, and Tsallis models are approximately the same, with free surface data (y/P = 0) and a data between the wall and the bed of the channel (y/P = 0.1) which are in outside of the CB.

This indicates that the accuracy of these three models is lower in estimating shear stress at the free surface and the boundary of the wall and bed than other wet points. Although the Renyi entropy model has more data within the CB, its WCB is much larger than the three Shannon, Shannon PL, and Tsallis models. Comparing the uncertainty of the Renyi and $\rho gRs$, it can be said that the two models have approximately the same WCB and $N_{in}$, but the certainty of the Renyi model at the intersection of the wall and bed and the certainty of the $\rho gRs$ at the free surface are lower than other wet points.

**Figure 6.** 95% *CB* for uncertainty analyzing presented models in shear stress prediction by HBMES-1 method in height ratios (**a**) t/D = 0, h+t/D = 0.333; and (**b**) t/D = 0.25; h+t/D = 0.333).

*3.3. Comparison of the uncertainty of four entropy models using HBMES-2 method*

Through drawing 95% CB, it was found that all models had sufficient certainty in predicting shear stress in circular and circular with flatbed channels, and the strengths and weaknesses of each model were determined. However, it is challenging and almost impossible to provide a classification and opinion on the final degree of uncertainty of each model.

Therefore, the results of the HBEMS-2 uncertainty method are presented below to compare the uncertainty of the models. In this method, by drawing the minimum confidence bound (*OBC*) all measured values are dropped in the band, and the $N_{in}$ statistic is equal to 100% and eliminated to check the uncertainty. Also, the *FREE* value introduced in previous studies [36,42,45,46] is optimized (*FREE$_{opt}$*) and according to Equations (22) and (23) are obtained. The transfer factor of Box-Cox function derived from the best SS was used to evaluate the uncertainty of shear stress predictor models.

The introduced statistics with HBMES-2 were obtained for 23 of sample for all entropy models, some of which are presented in Table 2. In the first column of this table, the *OCB* statistic shows the minimum CB that represents all measured data. The lower the *OCB* value, the higher the model's certainty in the shear stress prediction. The *FREE$_{opt}$* statistic, which is the total distance of the measured data of the *OCB*, represents the width of *OCB*. If this value is lower for a model, the certainty of this model is higher.

As mentioned before, the *OCB* criterion can be considered as a quantitative criterion for determining the certainty of a model. Also, *FREE$_{opt}$* is a qualitative criterion to evaluate the uncertainty of models. The third column of Table 2 is the value that shows the combined effect of *OCB* and *FREE$_{opt}$*, which is considered as the main criterion in this study as (*FOCB*) (*FREE$_{opt}$* and *OCB*).

Therefore, it is sufficient to find only the *FOCB* statistic for evaluating the uncertainty of shear stress predictor models. As can be seen in the last column of Table 2, the overall uncertainties of all models decrease with an increasing amount of water and sediment in the channel bed (h+t). In all samples, the *FOCB* statistic has the lowest value for the Shannon PL entropy and the highest value for the Renyi entropy, as a result, among the proposed models, Shannon PL models showed the highest certainty, and the Renyi has the lowest certainty.

The *FOCB* value for the Renyi entropy is more than several times of the other models, and this indicates a much lower certainty of this model than the other models in shear stress prediction. In the three samples 1, 2, and 11, the entropies of Shannon, $\rho gRs$, and Tsallis have the lowest amount of the *FOCB* and the highest certainty, respectively. In samples 8 and 11, the values of the *FOCB* for the three entropies Shannon, $\rho gRs$, and Tsallis are also close together, indicating an almost identical degree of certainty for these models.

To illustrate the uncertainty statistics presented in Table 3, the *OCB* for two samples 1 and 11 is shown in Figure 7. Figure 7 shows the minimum confidence bound that covers all measured shear stress, *OCB*, for four entropy models and $\rho gRs$ model. The Figure 7a corresponds to Sample 1 for a circular channel ($t/D = 0$, $h+t/D = 0.333$) and the Figure 7b also relates to sample 11 for a circular with flatbed channel ($t/D = 0.25$, $h+t/D = 0.333$).

Each model, which has a higher *OCB* width, shows higher uncertainty in the prediction of shear stress. The width of *OCB* of the three entropy models of Tsallis, Shannon PL, and Shannon is very close and contains a very small region, while the Renyi entropy model has larger *OCB* region compared with other three models.

**Table 3.** Statistical indexes based on HBMES-2 uncertainty method for four entropy models for shear stress prediction.

In Figure 7a the *OCB* widths for the two Shannon PL and Shannon models are perfectly within the range of the *OCB* plotted for the $\rho gRs$ model, but the *OCB* plotted for the Tsallis entropy is slightly wider than the *OCB* for the $\rho gRs$ model. Therefore, the certainty of the Shannon PL, Shannon, $\rho gRs$, Tsallis, and Renyi, respectively, is higher in predicting shear stress in this sample. The results of the *FOCB* values also confirm this classification.

In Table 3, sample 1 ($t/D = 0$), the *OCB* values for three Shannon PL, Shannon, and $\rho gRs$ models are close to each other and are less than the *OCB* values of two Tsallis and Renyi models. Despite the proximity of the *OCB* values for the two Tsallis and Renyi models, the width of *OCB* for the Renyi model is much higher than Tsallis model; the $FREE_{opt}$ value of the Renyi model is much larger than the other four models, which is shown in Table 3. In Figure 7b, although the *OCB* values are very close to each other for all models, the width of *OCB*, in bed and wall of channel, for the Renyi entropy model is much larger than the four models. This issue is very clear in the (*FOCB*) statistics in Table 3 for all samples.

**Figure 7.** *OCB* for analyzing presented models in shear stress prediction by HBMES-2 uncertainty method in height ratios (**a**) t/D=0, h+t/D=0.333; and (**b**) t/D=0.25; h+t/D=0.333).

In Table 4, the values of this statistic are given for all models for the studied cases. These values show that the certainty of all models to calculating shear stress in a circular channel is much higher than in circular with flatbed channel. Except for the Renyi entropy with identical almost values for *FOCB*, it indicates the same uncertainty in the prediction of shear stress values. In Table 4, the values of *FOCB* for the Renyi entropy in both circular and circular with flatbed channels are much higher than in the other models. Given that the $\rho gRs$ model has been considered as a basic model in this study for comparison with entropy models, it should be said that the Renyi entropy model is not a good model for predicting shear stress values. In a circular channel, the entropies of Shannonn PL, $\rho gRs$, Shannon, Tsallis have more certainty in predicting shear stress.

The models are more certain for Shannon, Shannon PL, $\rho gRs$, and Tsallis models to predicting shear stress in circular with flatbed channel, respectively. Finally, considering the number of samples presented in Table 1, the average values in Table 4 are given for the entire circular channel. The average values indicate that Shannon, Shannon PL, $\rho gRs$ and Tsallis models have the most certainty to estimating shear stress in the circular channels, respectively.

**Table 4.** The values of *FOCB* in circular and circular with flatbed channels for all entropy models to predicting shear stress distribution.

## 4. Conclusion

In this study, we present the uncertainty of four popular entropy models, including Shannon, Shannon PL, Tsallis, and Renyi have been analyzed for calculating shear stress in circular channels. The uncertainty analysis method of Kazemian-Kale-Kale et al. [46] based on the Bayesian Monte-Carlo technique was employed and named in this study as HBMES-1. However, using the HBMES-1 method along with the $N_{in}$ statistic required three additional statistics ($F_P$, $F_N$, and *FREE*) to compare the uncertainty of the model, it was not feasible to rank the uncertainty of several models. For this reason, in the proposed new HBMES-2 method, the minimum CB that covers all measured data and the *OCB* were obtained, and a new statistic called *FOCB* was introduced to evaluate the uncertainty. The *FOCB* indicates the degree of uncertainty as both a qualitative and a quantitative index. Calibrations for both HBMES-1 and HBMES-2 methods were performed in different sample sizes and under different hydraulic conditions. Based on the percentage of measured data within the confidence bound ($N_{in}$), the best SS (sample size) for each entropy model was selected. At the calibration stage, the mean and standard deviation changes of the error distribution of normalized shear stress data ($\delta_\varepsilon$ and $\mu_\varepsilon$) of all four entropy models in different SS were investigated. The lower $\delta_\varepsilon$ values, the distribution of the error is approaching to the Gaussian distribution, and the validity of the obtained uncertainty results is more reliable. The values $\delta_\varepsilon$ for the entropies of Shannon, Shannon PL, Tsallis, and Renyi, were 0.09, 0.06, 0.08, and 0.21, respectively. The absolute values of $\mu_\varepsilon$ for the entropies of Shannon PL, Tsallis, Shannon, and Renyi, were 0.001, 0.003, 0.055, and 0.114, respectively. Applying the Box-Cox function transfer factor in the best SS selected from the model calibration, the final evaluation was performed using two uncertainty methods. In the HBMES-1 method, it was found that all four entropy models with a percentage of measured data within 95% CB higher than 93% along with the $\rho gRs$ conventional model have high certainty in predicting shear stress in circular channels. According to the results of HBMES-2 method in a circular channel, the entropies of Shannon PL, $\rho gRs$, Shannon, Tsallis, and Renyi with the *FOCB* values were 1.339, 2.026, 2.432, 2.961, and 58.457, respectively. In the circular flatbed channel, the entropies of Shannon PL, Shannon, $\rho gRs$, Tsallis, and Renyi, the *FOCB* values were 10.118, 11.591, 17.115, 17.407, and 57.565, respectively. According to the mean results of *FOCB* in different hydraulic conditions, it was generally found that the Shannon PL, Shannon, $\rho gRs$, Tsallis, and Renyi models have the highest certainty in shear stress of circular channels with *FOCB* values of 8.781, 9.808, 14.491, 14.895, and 57.726, respectively. These results show that the three Shannon, Shannon PL, and Tsallis entropy models, along with the $\rho gRs$ conventional model, have high certainty in shear stress prediction, whereas the Renyi entropy model has the least certainty in predicting shear stress in the circular channels.

# References

[1] Kleijwegt, R.A. On sediment transport in circular sewers with non-cohesive deposits. Ph.D. Thesis, Delft University of Technology, Delft, Netherlands, 1992.

[2] Knight, D.W.; Sterling, M. Boundary shear in circular pipes running partially full. Journal of Hydraulic Engineering **2000**, *126* (4), 263–275.

[3] Perrusquia, G. Bedload Transport in Storm Sewers. Stream Traction in Pipe Channels. Ph.D. Thesis, Chalmers University of Technology, Gothenburg, Sweden, 1991.

[4] Berlamont, J.E.; Trouw, K.; Luyckx, G. Shear stress distribution in partially filled pipes. Journal of Hydraulic Engineering **2003**, *129* (9), 697–705.

[5] Bares, V.; Jirak, J.; Pollert, J. Bottom shear stress in unsteady sewer flow. Water Science & Technology **2006**, *54* (6–7), 93–100.

[6] Yu, G.; Tan, S.K. Estimation of boundary shear stress distribution in open channels using flownet. Journal of Hydraulic Research **2007**, *45*, 486–496.

[7] Khozani, Z.S.; Bonakdari, H.; Zaji, A.H. Estimating the shear stress distribution in circular channels based on the randomized neural networks technique. Applied Soft Computing **2017**, *58*, 441–448.

[8] Khozani, Z.S.; Bonakdari, H.; Ebtehaj, I. An analysis of shear stress distribution in circular channels with sediment deposition based on Gene Expression Programming. International journal of Scientific research & communications **2017**, *32*(4), 575-584.

[9] Farina, G.; Alvisi, S.; Franchini, M.; Moramarco, T. Three methods for estimating the entropy parameter M based on a decreasing number of velocity measurements in a river cross-section. Entropy **2014**, *16*(5), 2512-2529.

[10] Kumbhakar, M.; Kundu, S.; Ghoshal, K. Hindered settling velocity in particle-fluid mixture: a theoretical study using the entropy concept. Journal of Hydraulic Engineering **2017**, *143*(11), 06017019.

[11] James, R.D.; Nota, A.; Velázquez, J.J. Self-similar profiles for homoenergetic solutions of the Boltzmann equation: particle velocity distribution and entropy. Archive for Rational Mechanics and Analysis **2019**, *231*(2), 787-843.

[12] Gholami, A.; Bonakdari, H.; Mohammadian, M.; Zaji, A.H.; Gharabaghi, B. Assessment of geomorphological bank evolution of the alluvial threshold rivers based on entropy concept parameters. Hydrological Sciences Journal **2019a**, *64*(7), 856-872.

[13] Gholami, A.; Bonakdari, H.; Mohammadian, M. Enhanced formulation of the probability principle based on maximum entropy to design the bank profile of channels in geomorphic threshold. Stochastic Environmental Research and Risk Assessment. **2019b**, 1-22.

[14] Gholami, A.; Bonakdari, H.; Mohammadian, A. A method based on the Tsallis entropy for characterizing threshold channel bank profiles. Physica A: Statistical Mechanics and its Applications **2019c**, *526*, 121089.

[15] Bonakdari, H.; Tooshmalani, M.; Sheikh, Z. Predicting shear stress distribution in rectangular channels using entropy concept. International Journal of Engineering **2015**, *28*(3), 360-367.

[16]     Chiu, C.L. Entropy and probability concepts in hydraulics. Journal of Hydraulic Engineering **1987**, *113*, 583–599.

[17]     Sterling, M.; Knight, D.W. An attempt at using the entropy approach to predict the transverse distribution of boundary shear stress in open channel flow. Stochastic Environmental Research and Risk Assessment **2002**, *16*(2), 127–142.

[18]     Sheikh, Z.; Bonakdari, H. Prediction of boundary shear stress in circular and trapezoidal channels with entropy concept. Urban Water Journal **2015**, *13*(6), 629–636.

[19]     Ghoshal, K.; Kumbhakar, M.; Singh, V.P. Distribution of sediment concentration in debris flow using Rényi entropy. Physica A: Statistical Mechanics and its Applications **2019**, *521*, 267-281.

[20]     Tsallis, C. Possible generalization of Boltzmann–Gibbs statistics. Journal of Statistical Physics **1988**, *52*(1–2), 479–487.

[21]     Rényi, A. On measures of entropy and information. Proc. Fourth Berkeley Symp. on Math. Statist. and Prob., Vol. 1 (Univ. of Calif. Press, 1961), 547-561, 1961.

[22]     Tsallis, C.; Brigatti, E. Nonextensive statistical mechanics: A brief introduction. Continuum Mechanics and Thermodynamics **2004**, *16*(3), 223-235.

[23]     Tsallis, C. Introduction to nonextensive statistical mechanics: approaching a complex world. Springer-Verlag, New York, USA, 2009.

[24]     Bonakdari, H.; Sheikh, Z.; Tooshmalani. M. Comparison between Shannon and Tsallis entropies for prediction of shear stress distribution in open channels. Stochastic Environmental Research and Risk Assessment **2015**, *29*(1), 1–11.

[25]	Khozani, Z.S.; Bonakdari, H. Formulating the shear stress distribution in circular open channels based on the Renyi entropy. Physica A: Statistical Mechanics and its Applications **2018**, *490*, 114–126.

[26]	Arnaud, P., Cantet, P.; Odry, J. Uncertainties of flood frequency estimation approaches based on continuous simulation using data resampling. Journal of Hydrology **2017**, *554*, 360-369.

[27]	Baldassarre, G.D.; Montanari, A. Uncertainty in river discharge observations: a quantitative analysis. Hydrology and Earth System Sciences **2009**, *13*(6), 913-921.

[28]	Beven, K.; Lamb. R. The uncertainty cascade in model fusion. Geological Society, London, Special Publications **2017**, *408*(1), 255-266.

[29]	Gabellani, S.; Boni, G.; Ferraris, L.; Hardenberg, J.V.; Provenzale, A. Propagation of uncertainty from rainfall to runoff: A case study with a stochastic rainfall generator. Advances in Water Resources **2007**, *30*(10), 2061-2071.

[30]	Gholami, A.; Bonakdari, H.; Ebtehaj, I.; Mohammadian, M.; Gharabaghi, B.; Khodashenas, S.R. Uncertainty analysis of intelligent model of hybrid genetic algorithm and particle swarm optimization with ANFIS to predict threshold bank profile shape based on digital laser approach sensing. Measurement **2018a**, *121*, 294–303.

[31]	Gholami, A.; Bonakdari, H.; Zaji, A.H.; Akhtari, A.A. A comparison of artificial intelligence-based classification techniques in predicting flow variables in sharp curved channels. Engineering with Computers **2019d**, 1-30.

[32]	Ghorbani, M.A.; Zadeh, H.A.; Isazadeh, M.; Terzi, O. A comparative study of artificial neural network (MLP, RBF) and support vector machine models for river flow prediction. Environmental Earth Sciences **2016**, *75*(6), 476.

[33] Hamilton, A.S.; Moore, R.D. Quantifying uncertainty in streamflow records. Canadian Water Resources Journal **2012**, *37*(1), 3-21.

[34] Iskra, I.; Droste, R. Parameter uncertainty of a watershed model. Canadian Water Resources Journal **2008**, *33*(1), 5-22.

[35] Iyer, V.A.; Woodmansee, M.A. Uncertainty analysis of laser-doppler-velocimetry measurements fin a swirling flowfield. The American institute of Aeronautics and Astronautics **2005**, *43*(3), 512-519.

[36] Misirli, F.; Gupta, H.V.; Sorooshian, S.; Thiemann, M. Bayesian recursive estimation of parameter and output uncertainty for watershed models. Calibration of Watershed Models **2003**, 113–124.

[37] Montanari, A.; Brath, A. A stochastic approach for assessing the uncertainty of rainfall-runoff simulations. Water Resources Research **2004**, *40*(1).

[38] Sattar, A.M.; Gharabaghi, B; McBean, E. Prediction of Timing of Watermain Failure Using Gene Expression Models. Water Resources Management **2016**, *30*(5), 1635-1651.

[39] Thompson, J.; Sattar, A.M.; Gharabaghi, B.; Richard, W. Event based total suspended sediment particle size distribution model. Journal of Hydrology **2016**, *536*, 236-246.

[40] Watt, W.E.; Paine, J.D. Flood risk mapping in Canada: 1. Uncertainty considerations. Canadian Water Resources Journal **1992**, *17*(2), 129-138.

[41] Lamb, R.; Beven, K.; Myrabø, S. Use of spatially distributed water table observations to constrain uncertainty in a rainfall–runoff model. Advances in water resources **1998**, *22*(4), 305-317.

[42]     Corato, G.; Melone, F.; Moramarco, T.; Singh, V.P. Uncertainty analysis of flow velocity estimation by a simplified entropy model. Hydrology Processes **2014**, *28*(3), 581–590.

[43]     Moramarco, T.; Saltalippi, C.; Singh, V.P. Velocity profiles assessment in natural channels during high floods. Hydrology Research **2011**, *42*(2-3), 162-170.

[44]     Rincon, R.A.C. Evaluation of uncertainty in hydrodynamic modeling. PhD. Thesis, Mississippi State University, USA, 2013.

[45]     Kazemian-Kale-Kale, A.; Bonakdari, H.; Gholami, A.; Khozani, Z.S.; Akhtari, A.A.; Gharabaghi, B. Uncertainty analysis of shear stress estimation in circular channels by Tsallis entropy. Physica A: Statistical Mechanics and its Applications **2018**, *510*, 558-576.

[46]     Kazemian-Kale-Kale, A.; Bonakdari, H.; Gholami, A.; Gharabaghi, B. The uncertainty of the Shannon entropy model for shear stress distribution in circular channels. International Journal of Sediment Research **2020**, *35*(1), 57-68.

[47]     Knight, D.W.; Yuen, K.W.H.; Alhamid, A.A.I. Boundary shear stress distributions in open channel flow. Physical Mechanisms of mixing and Transport in the Environment, John Wiley, New York, USA, 1993.

[48]     Sterling, M. A Study of Boundary Shear Stress, Flow Resistance and the Free Overfall in Open Channels with a Circular Cross-Section. Ph.D. dissertation, University of Birmingham, Birmingham, UK, 1998.

[49]     Abbaspour, K.C.; Johnson, C.A.; Van Genuchten, M.T. Estimating uncertain flow and transport parameters using a sequential uncertainty fitting procedure. Vadose Zone Journal **2004**, *3*(4), 1340-1352.

[50]     Abbaspour, K.C.; Yang, J.; Maximov, I.; Siber, R.; Bogner, K.; Mieleitner, J.; Srinivasan, R. Modelling hydrology and water quality in the pre-alpine/alpine Thur watershed using SWAT. Journal of hydrology **2007**, *333*(2-4), 413-430.

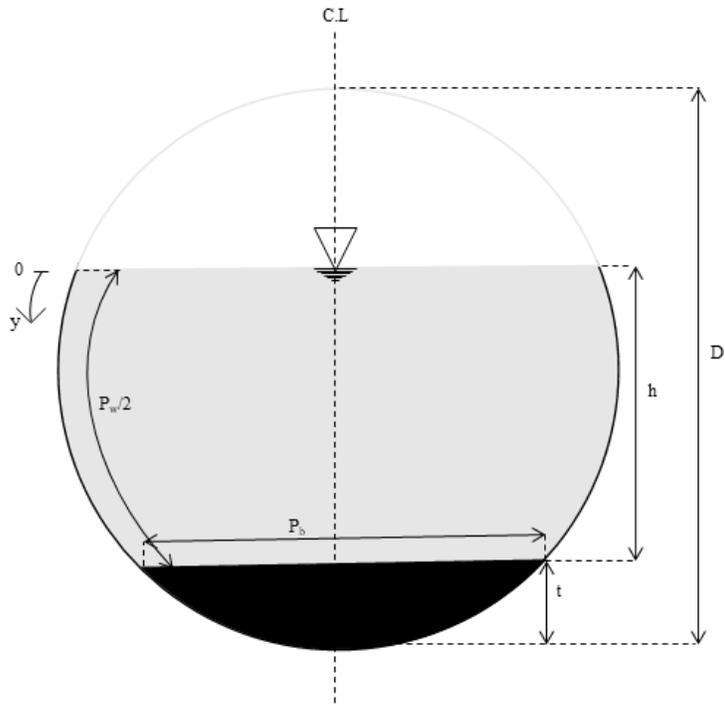

Figure 1. Cross-section of the circular with the flatbed channel and its notation

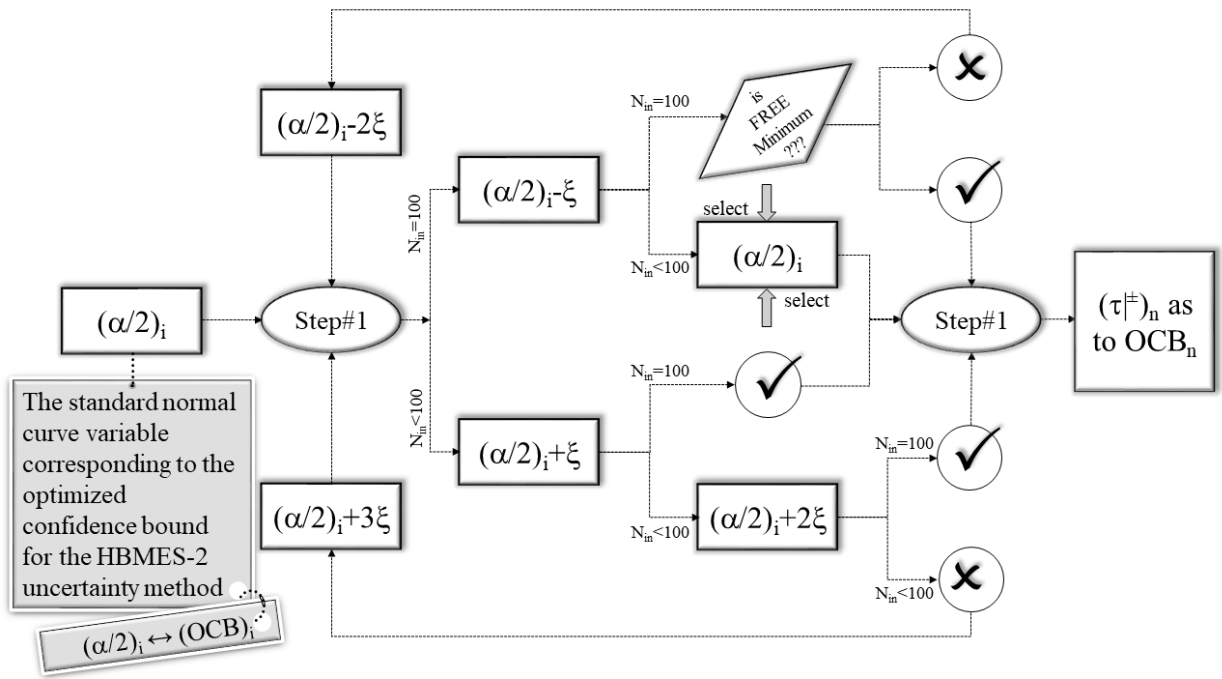

Figure 2. Process of assessment of the $OCB_n$ (minimum confidence bound that covers all measured data) in HBMES-2 uncertainty method

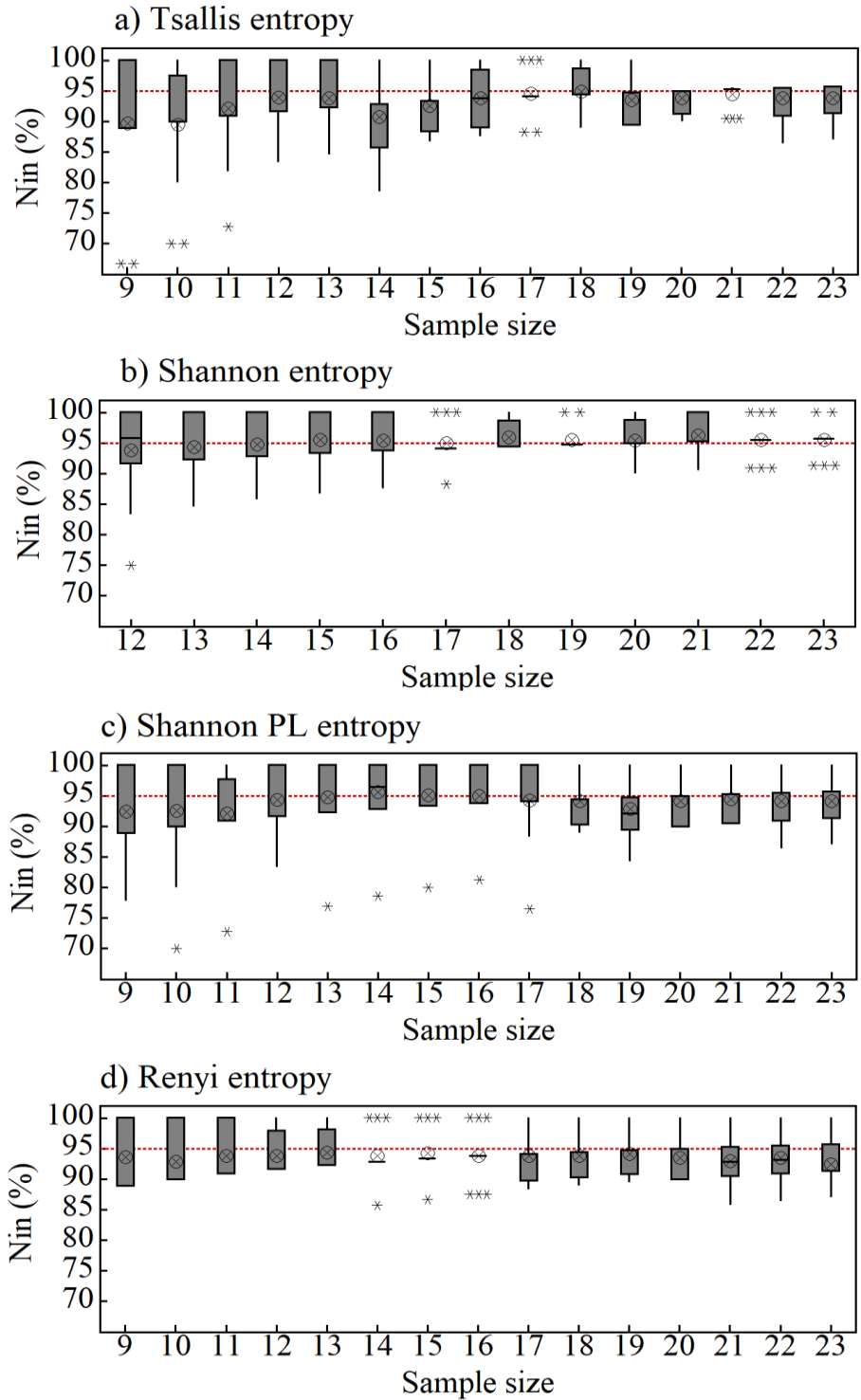

Figure 3. The changes in the percentage of measured values within the confidence bound (Nin) in different SS values for all entropy models

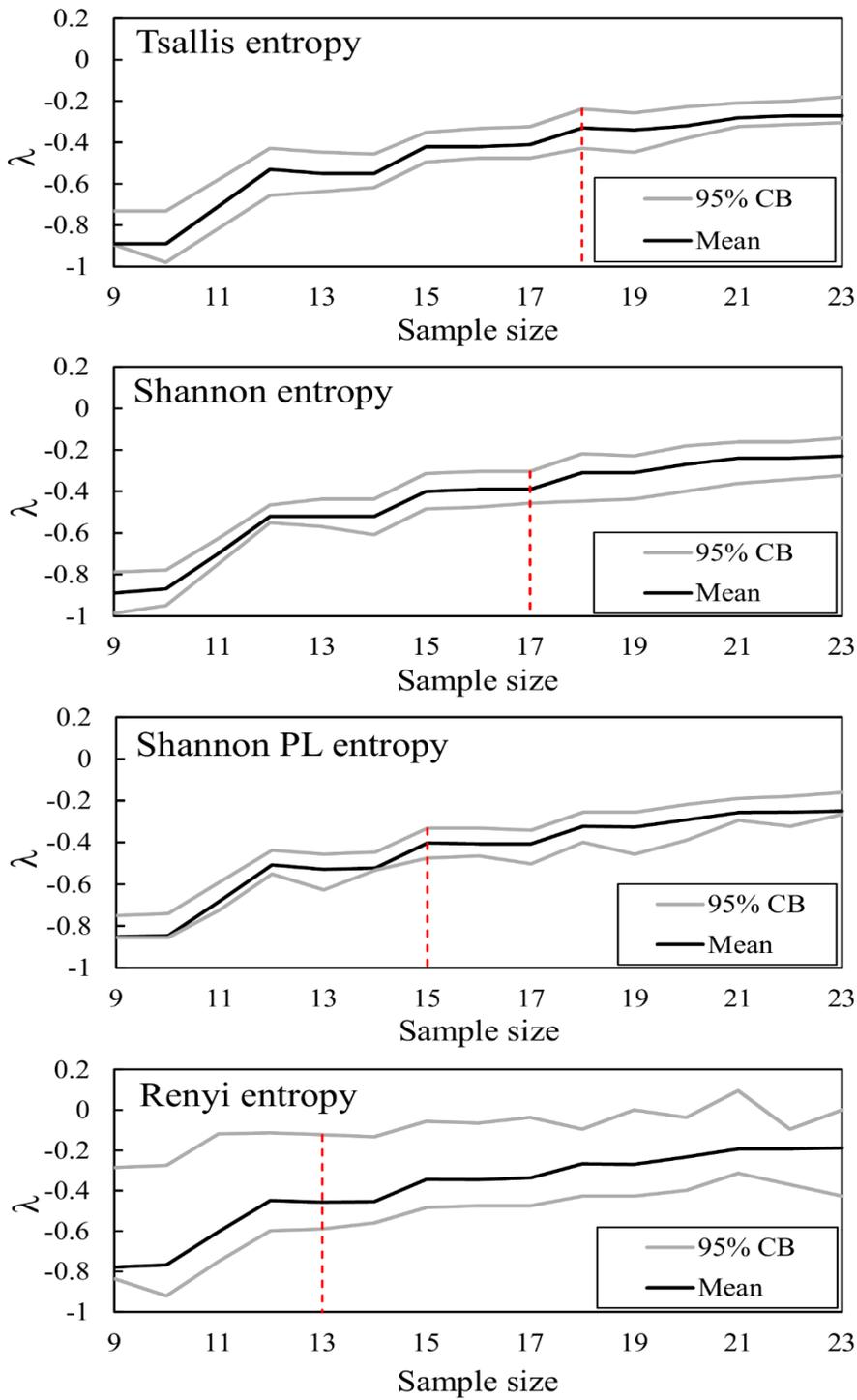

Figure 4. Changes of transfer factor ($\lambda$) values in different sample sizes for all entropy models

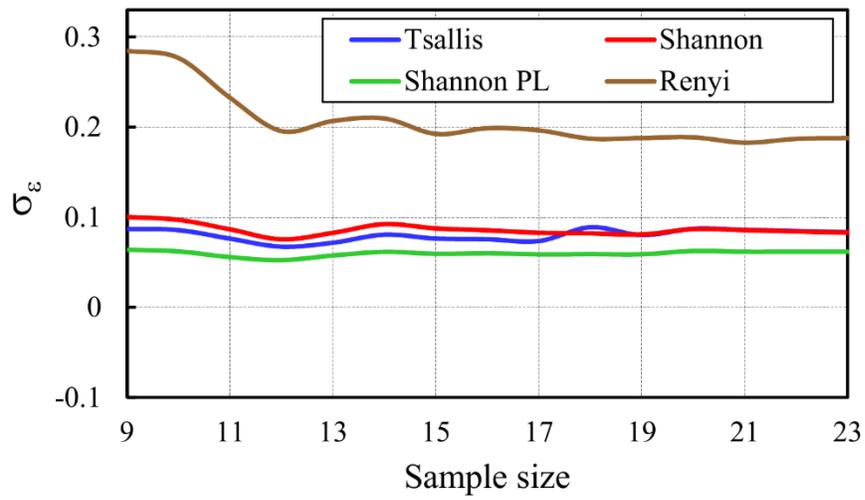
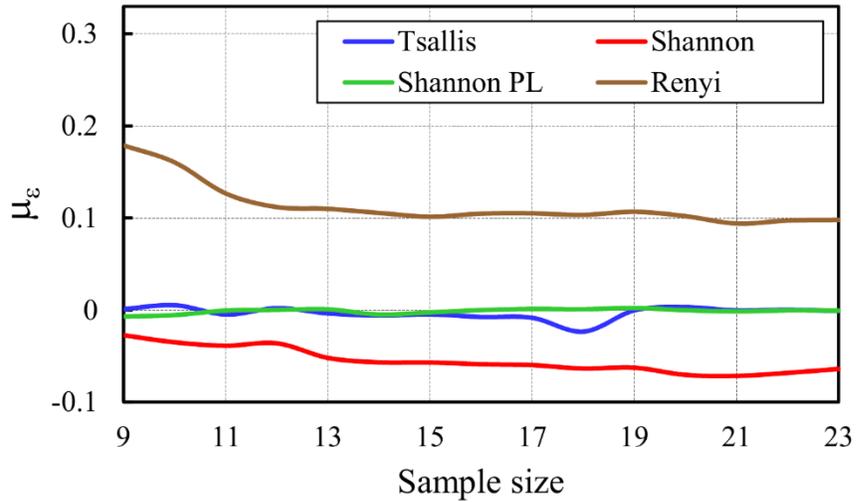

Figure 5. The mean and standard deviation of the Gaussian error distribution; δε and με, respectively; versus sample size in all entropy models

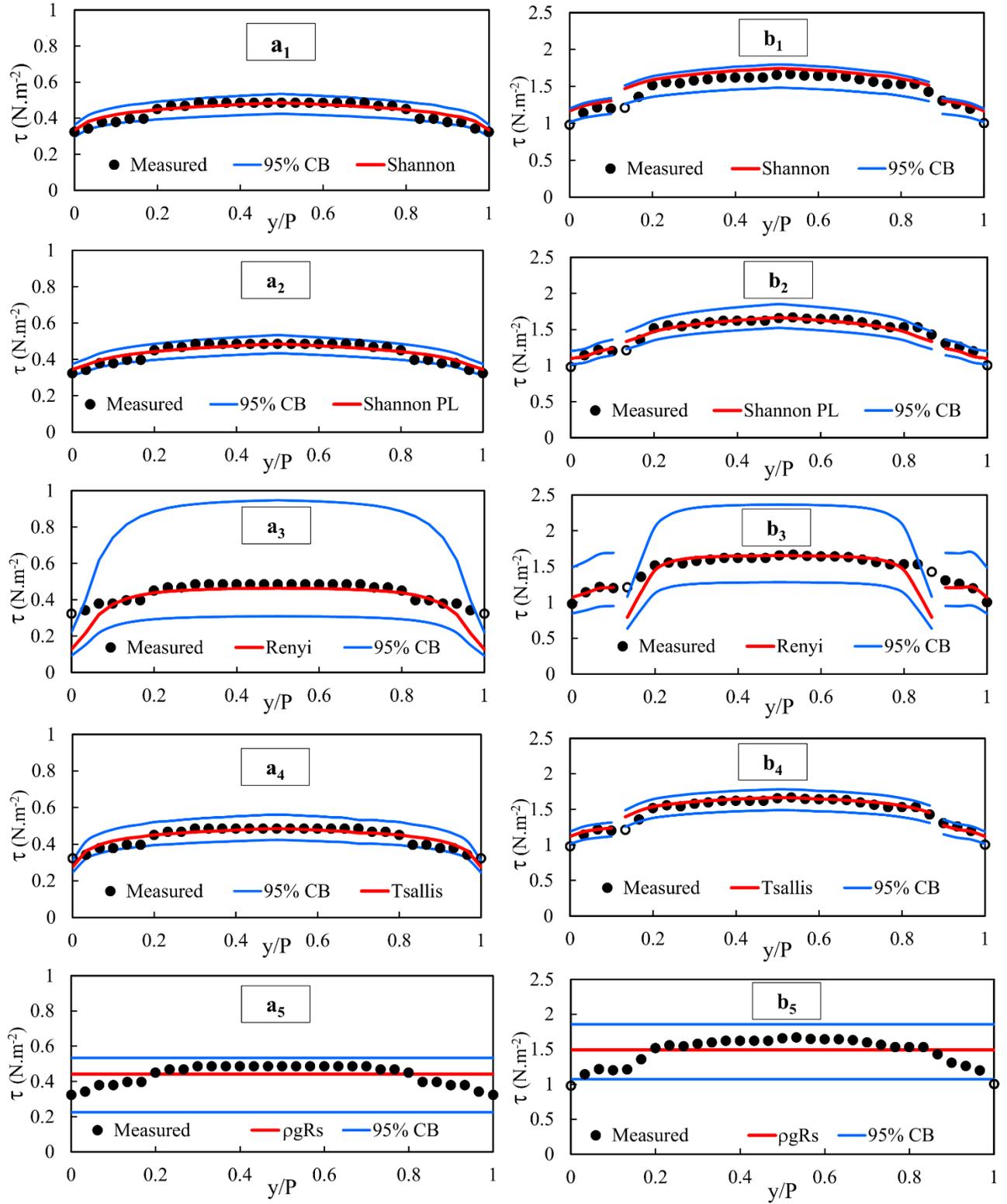

Figure 6. 95% *CB* for uncertainty analyzing presented models in shear stress prediction by HBMES-1 method in height ratios a) t/D = 0, h+t/D = 0.333; and b) t/D = 0.25; h+t/D = 0.333)

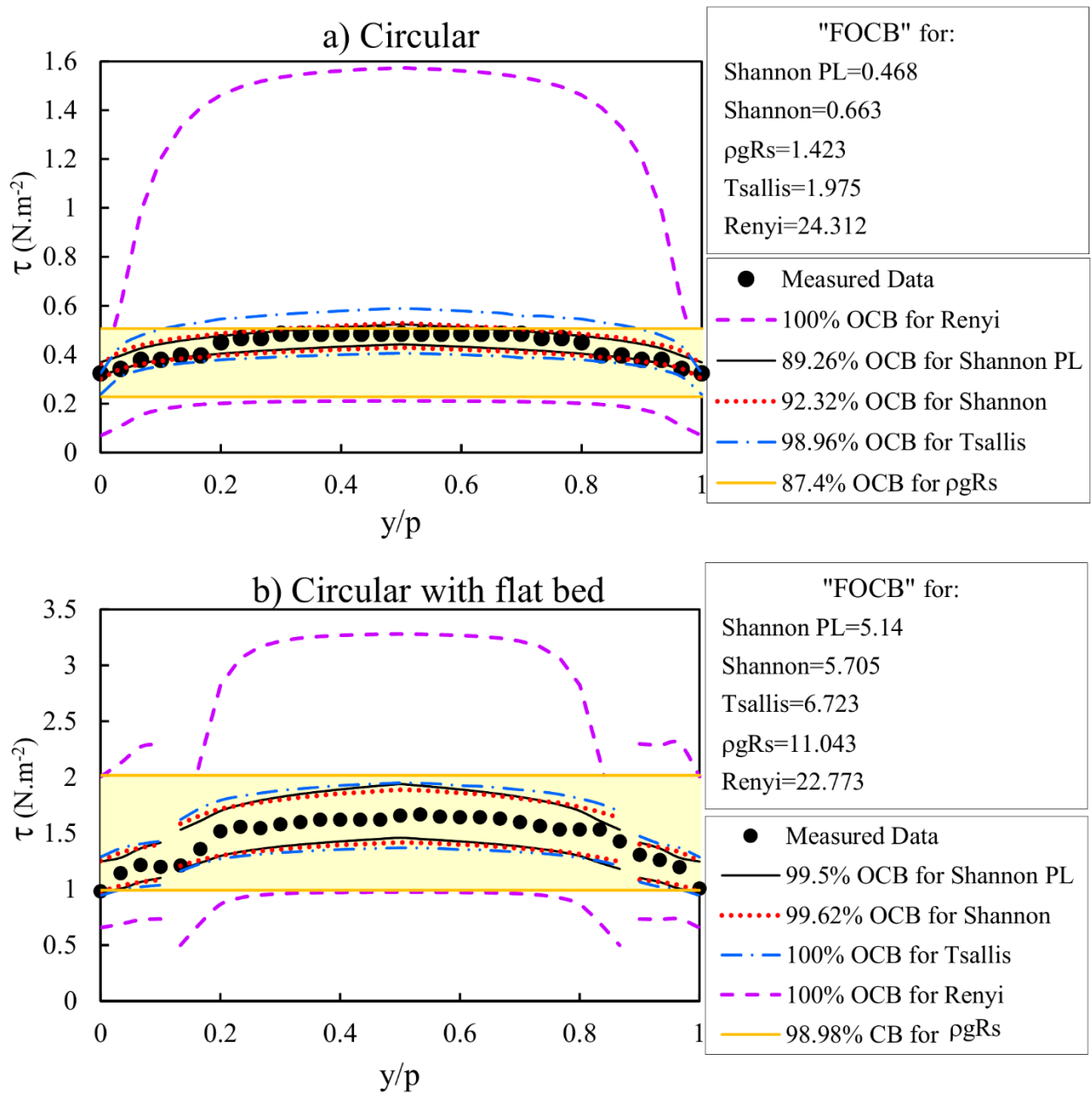

Figure 7. *OCB* for analyzing presented models in shear stress prediction by HBMES-2 uncertainty method in height ratios a) t/D=0, h+t/D=0.333; and b) t/D=0.25; h+t/D=0.333)

Table 1. Summary of the main hydraulic parameters in the circular channel with and without sediment

| Sample | Section | t/D | h+t/D | $S_0 \times 10^3$ | Fr | Q(l.s$^{-1}$) |
|---|---|---|---|---|---|---|
| 1 | Circular | 0 | 0.333 | 1.00 | 0.516 | 5.36 |
| 2 | | | 0.506 | 1.00 | 0.505 | 11.7 |
| 3 | | | 0.666 | 1.00 | 0.441 | 17.3 |
| 4 | | | 0.826 | 1.00 | 0.375 | 22.9 |
| 5 | Circular with flatbed | 0.25 | 0.332 | 1.96 | 0.671 | 1.32 |
| 6 | | | 0.499 | 1.96 | 0.748 | 8 |
| 7 | | | 0.398 | 1.96 | 0.656 | 3.3 |
| 8 | | | 0.666 | 1.96 | 0.68 | 16.5 |
| 9 | | | 0.755 | 1.96 | 0.663 | 22.1 |
| 10 | | | 0.795 | 1.96 | 0.626 | 23.8 |
| 11 | | | 0.333 | 8.62 | 1.71 | 3.39 |
| 12 | | | 0.499 | 8.62 | 1.7 | 18.2 |
| 13 | | | 0.666 | 8.62 | 1.59 | 38.9 |
| 14 | Circular with flat bed | 0.332 | 0.499 | 2.00 | 0.718 | 4.4 |
| 15 | | | 0.666 | 2.00 | 0.685 | 12.2 |
| 16 | | | 0.75 | 2.00 | 0.669 | 17 |
| 17 | | | 0.8 | 2.00 | 0.721 | 22.1 |
| 18 | | | 0.499 | 2.00 | 1.96 | 12 |
| 19 | Circular with flatbed | 0.5 | 0.666 | 9.00 | 1.4 | 8.4 |
| 20 | | | 0.75 | 9.00 | 1.42 | 16 |

| | | | | | |
|---|---|---|---|---|---|
| 21 | | 0.8 | 9.00 | 1.33 | 20 |
| 22 | Circular with | 0.75 | 8.80 | 1.44 | 3.09 |
| | 0.664 | | | | |
| 23 | flat bed | 0.8 | 8.80 | 1.55 | 4.93 |

Table 2 Statistical indexes based on HBMES-1 uncertainty method in shear stress prediction by different entropy models and conventional $gRs$ model

|  | Models | $N_{in}$ | $F_P$ | $F_N$ | FREE |
|---|---|---|---|---|---|
| Entropy | Shannon | 94.81 | 6.521 | 0.096 | 6.617 |
|  | Shannon PL | 93.41 | 6.018 | 0.107 | 6.125 |
|  | Tsallis | 92.43 | 8.525 | 0.239 | 8.764 |
|  | Renyi | 91.58 | 26.041 | 0.658 | 26.699 |
| Conventional | $gRs$ | 85.41 | 8.124 | 1.715 | 9.839 |

Table 3. Statistical indexes based on HBMES-2 uncertainty method for four entropy models for shear stress prediction

| Samples | Models | OCB | FREE$_{opt}$ | FOCB |
|---|---|---|---|---|
| 1 | Shannon PL | 89.26 | 0.525 | 0.469 |
| | Tsallis | 98.96 | 1.995 | 1.974 |
| | Shannon | 92.32 | 0.718 | 0.663 |
| | Renyi | 100 | 24.312 | 24.312 |
| | gRs | 87.4 | 1.628 | 1.423 |
| 2 | Shannon PL | 98.44 | 0.768 | 0.756 |
| | Tsallis | 100 | 4.057 | 4.057 |
| | Shannon | 95.86 | 0.975 | 0.935 |
| | Renyi | 100 | 60.569 | 60.569 |
| | gRs | 93.14 | 1.480 | 1.379 |
| 8 | Shannon PL | 94.76 | 1.927 | 1.826 |
| | Tsallis | 97.6 | 1.953 | 1.906 |
| | Shannon | 96.06 | 2.931 | 2.815 |
| | Renyi | 99.02 | 29.049 | 28.764 |
| | gRs | 99.14 | 4.920 | 4.878 |
| 11 | Shannon PL | 99.5 | 5.166 | 5.14 |
| | Tsallis | 100 | 6.723 | 6.723 |
| | Shannon | 99.62 | 5.727 | 5.705 |
| | Renyi | 100 | 22.773 | 22.773 |
| | gRs | 98.98 | 11.157 | 11.043 |
| | Shannon PL | 100 | 17.426 | 17.426 |
| | Tsallis | 100 | 22.231 | 22.231 |
| 18 | Shannon | 100 | 16.049 | 16.049 |
| | Renyi | 100 | 41.814 | 41.814 |
| | ☐gRs | 99.56 | 21.233 | 21.139 |

Table 4. The values of *FOCB* in circular and circular with flatbed channels for all entropy models to predicting shear stress distribution

| Section | FOCB | | | | |
| --- | --- | --- | --- | --- | --- |
| | Shannon PL | Shannon | Tsallis | Renyi | ☐gRs |
| Circular | 1.339 | 2.432 | 2.961 | 58.457 | 2.026 |
| Circular with flat bed | 11.591 | 10.118 | 17.407 | 57.569 | 17.115 |
| Average | 9.808 | 8.781 | 14.895 | 57.726 | 14.491 |